\documentclass[letterpaper, 10 pt, conference]{ieeeconf}  

\IEEEoverridecommandlockouts                              

\usepackage{graphics} 
\usepackage{epsfig} 
\usepackage{mathptmx} 
\usepackage{times} 
\usepackage{amsmath,amssymb}
\usepackage{subfig}
\usepackage[T1]{fontenc} 
\usepackage{cite}
\usepackage{booktabs}

\usepackage{algorithmic}
\usepackage{setspace}
\usepackage{graphicx}
\usepackage{textcomp}
\usepackage{xcolor}
\usepackage[T1]{fontenc}
\usepackage{aecompl}
\usepackage{multirow}
\usepackage{threeparttable}
\usepackage{tabularx}
\usepackage{mathtools}
\usepackage[super]{nth}
\usepackage[free-standing-units=true]{siunitx}
\usepackage[colorinlistoftodos,prependcaption,textsize=tiny]{todonotes}
\usepackage{balance}

\usepackage[ruled,vlined]{algorithm2e}
\newcommand{\Rom}[1]{\uppercase\expandafter{\romannumeral #1\relax}}
\newcommand{\rom}[1]{\expandafter{\romannumeral #1\relax}}

\usepackage{array}
\newcommand{\PreserveBackslash}[1]{\let\temp=\\#1\let\\=\temp}
\newcolumntype{C}[1]{>{\PreserveBackslash\centering}p{#1}}
\newcolumntype{R}[1]{>{\PreserveBackslash\raggedleft}p{#1}}
\newcolumntype{L}[1]{>{\PreserveBackslash\raggedright}p{#1}}

\newcommand{\fref}[1]{Fig. \ref{#1}}
\newcommand{\sref}[1]{Section \ref{#1}}
\newcommand{\tref}[1]{Table \ref{#1}}

\usepackage{mathptmx,mathtools}
\DeclareFontFamily{U} {cmex}{}
\DeclareFontShape{U}{cmex}{m}{n}{
  <-6> cmex5
  <6-7> cmex6
  <7-8> cmex7
  <8-9> cmex8
  <9-10> cmex9
  <10-12> cmex10
  <12-> cmex12}{}
\DeclareSymbolFont{Xcmex} {U} {cmex}{m}{n}
\DeclareMathSymbol{\Xsum}{\mathop}{Xcmex}{80}

\makeatletter
\let\NAT@parse\undefined
\makeatother

\usepackage[colorlinks]{hyperref}
  \hypersetup{
    citecolor=blue,
    linkcolor=blue,   
    urlcolor=blue}

\author{Xiao Lin$^{1}$, Yuhao Huang$^{2}$, Taimeng Fu$^{1}$, Xiaobin Xiong$^{2}$, and Chen Wang$^{1}$
\thanks{$^{1}$The Spatial AI \& Robotics (SAIR) Lab, Computer Science and Engineering, University at Buffalo, NY 14260, USA. Email: {\tt\small \{xiaol, taimengf, chenw\}@sairlab.org}}%
\thanks{$^{2}$The Wisconsin Expeditious Legged Locomotion (WELL) Lab, Mechanical Engineering, University of Wisconsin-Madison, WI 53706, USA. Email: {\tt\small \{yuhao.huang, xiaobin.xiong\}@wisc.edu}}%
\thanks{$^\dagger$The project page and a real-time video demonstration for iWalker are available at \url{https://sairlab.org/iwalker/}.}
}
\begin{document}

\title{\LARGE \bf
iWalker: Imperative Visual Planning for Walking Humanoid Robot
}

\thispagestyle{empty}
\pagestyle{empty}

\maketitle

\begin{abstract}
Humanoid robots, designed to operate in human-centric environments, serve as a fundamental platform for a broad range of tasks.
Although humanoid robots have been extensively studied for decades, a majority of existing humanoid robots still heavily rely on complex modular frameworks, leading to inflexibility and potential compounded errors from independent sensing, planning, and acting components.
In response, we propose an end-to-end humanoid sense-plan-act walking system, enabling vision-based obstacle avoidance and footstep planning for whole body balancing simultaneously. 
We designed two imperative learning (IL)-based bilevel optimizations for model-predictive step planning and whole body balancing, respectively, to achieve self-supervised learning for humanoid robot walking.
This enables the robot to learn from arbitrary unlabeled data, improving its adaptability and generalization capabilities. 
We refer to our method as iWalker and demonstrate its effectiveness in both simulated and real-world environments, representing a significant advancement toward autonomous humanoid robots.
\end{abstract}

\section{Introduction}

The demand for autonomous humanoid robots has surged in recent years, as they offer unique capabilities in human-centric environments \cite{tong2024advancements}. Unlike wheeled or tracked robots, humanoid robots can navigate complex spaces, climb stairs, and open doors, making them ideal for roles such as personal assistants \cite{mivseikis2020lio} and emergency responders \cite{chitikena2023robotics}. As urban areas grow more complex, the use of humanoid robots in these roles is expected to rise significantly \cite{tiddi2020robot}.
Autonomous sensing, planning, and control are keys to realizing the full potential of humanoid robots, enabling them to move efficiently and respond to their environment \cite{ahn2021versatile}.

Methods based on traditional models for humanoid robot walking often face unreliability and scalability limitations\cite{khadiv2020walking, gao2021autonomous, delfin2018humanoid, maier2013integrated, wellhausen2021rough}. These approaches typically require separate perception, planning, and control modules, as well as the construction of 3D or 2.5D maps \cite{li2023autonomous}.
However, such modular architectures introduce the risk of \textit{compounded errors} \cite{yang2023iplanner}, as inaccuracies in one module can accumulate through the pipeline. For instance, an imprecisely detected obstacle in the perception module can exacerbate planning errors, leading to even greater deviations in the control module. Furthermore, these methods often require extensive hyperparameter tuning to ensure the system functions effectively, adding to their complexity and limiting adaptability.


\begin{figure}[t]
\vspace{8pt}
\centering\includegraphics[width=0.48\textwidth]{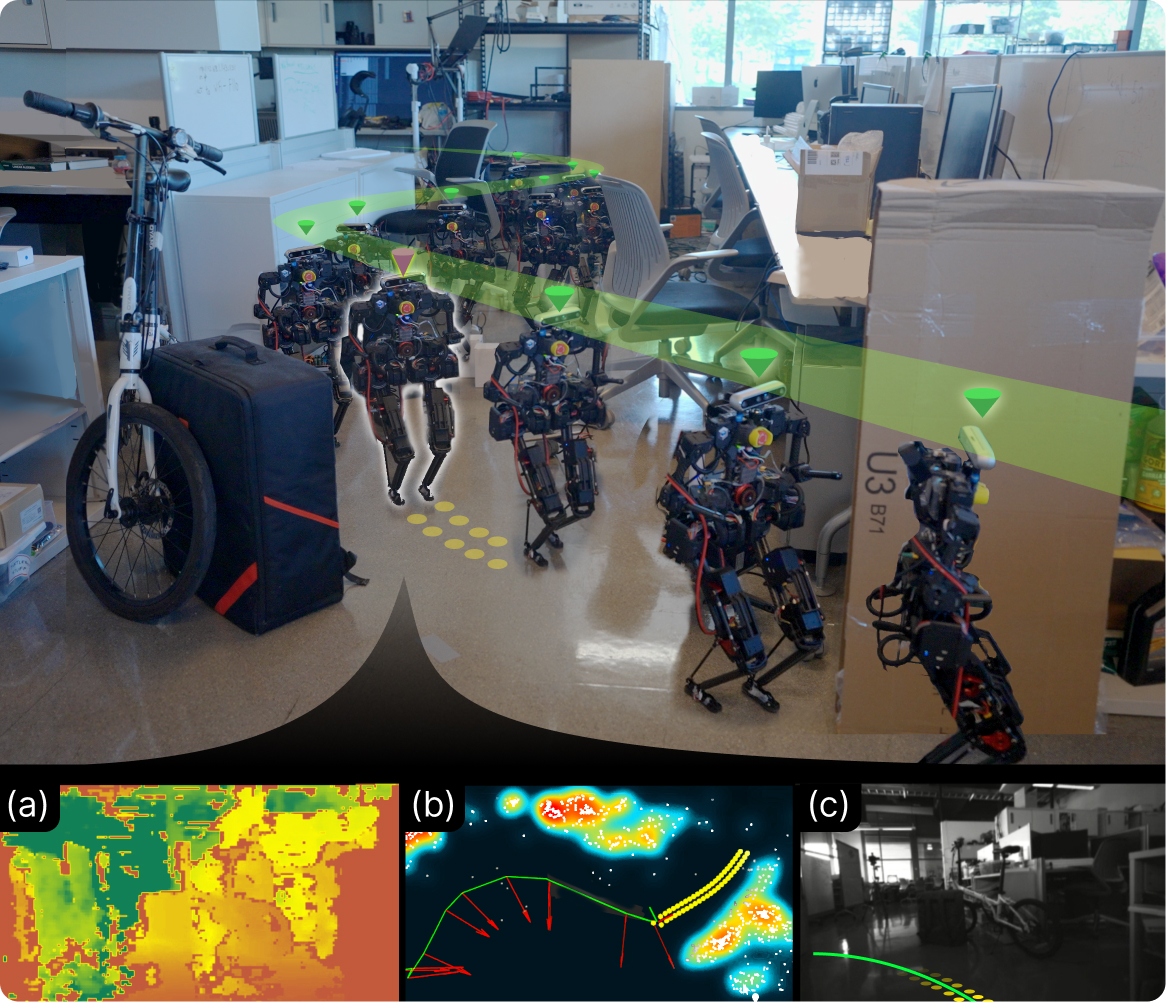}
\caption{A humanoid robot navigates autonomously through a messy office using iWalker. (a) is the input depth image; (b) is a visualization of the projected collision map; and (c) is a visualization of the planned path and footsteps.}
\vspace{-7pt}
\label{Fig. traj}
\end{figure}

Learning-based approaches such as reinforcement learning (RL) \cite{kaelbling1996reinforcement} offer promising alternatives for planning and locomotion \cite{li2021reinforcement, li2024reinforcementlearningversatiledynamic} but also need modular setup as their networks usually focus on a single task like planning or control only. In addition, these methods can suffer from sampling inefficiency, such as relying on stochastic sampling across vast action spaces or requiring extensive expert demonstrations \cite{bojarski2016end}. This limits their ability to generalize effectively in real-world applications. 
RL also faces challenges in generating precise low-level actions when given high-dimensional image inputs \cite{allshire2021laser}.
In addition, these approaches of optimizing and configuring perception, planning, and control modules separately can lead to suboptimal performance \cite{cao2023exploring,levine2016end}.

To eliminate the compounded errors from independent sensing, planning, and control components, there is an increasing trend to develop \emph{end-to-end systems} that seamlessly connect visual inputs to mid-level footstep planning, and whole body balancing, enabling accurate and robust task execution \cite{doukhi2021deep}. Such systems allow robots to adapt to complex environments, eliminating the need for task-specific reconfiguration or reengineering. We introduce a model predictive control (MPC)-based bilevel optimization (BLO) training pipeline inspired by imperative learning (IL) \cite{wang2024imperative,fu2024islam,chen2024iastar} to realize end-to-end sense-plan-act pipeline for walking humanoid robot. Our proposed system, iWalker, represents a framework that connects visual inputs directly to footstep control, bypassing the need for map construction and manual tuning. By integrating a collision map and MPC losses derived from any unlabeled depth data, iWalker leverages gradient-based optimization with embedded dynamics model, significantly enhancing its generalization capabilities. This approach simplifies traditional modular pipelines while offering improved performance, marking a significant step forward for walking autonomous humanoid robots.
In summary, the main contributions of this paper are:
\begin{itemize}
\item We designed a humanoid sense-plan-act walking system, enabling vision-based obstacle avoidance, footstep planning, and whole body balancing, simultaneously.
\item We designed a self-supervised and end-to-end training pipeline based on two MPC-based BLOs, which infuse dynamics constraints into a path planning network and a footstep planning network, respectively.
\item Through extensive experiments, we demonstrate that iWalker can adapt to various indoor environments, showing consistent performance of dynamics-feasibility in both simulations and on a real humanoid robot.
\end{itemize}

\section{Related Work}

\subsection{Path Planners}

Classic path planning methods have demonstrated significant success in navigating complex terrains. For example, Cao et al. \cite{cao2022autonomous}, Wellhausen and Hutter \cite{wellhausen2021rough}, and Yang et al. \cite{yang2021real} combined terrain analysis, mapping, and motion-primitive searches in real-world applications, including the DARPA SubT Challenge \cite{cao2023exploring}. These methods employ path search techniques such as lazyPRM \cite{kavraki2000path} and PRM* \cite{karaman2011sampling}, which have proven effective in various settings.
Recent advances shifted toward end-to-end planners that directly map sensory inputs to paths \cite{kim2022learning} or control actions \cite{sadat2020perceive}, addressing issues like increased latency and reduced robustness commonly found in modular systems. These planners often rely on imitation of expert demonstrations \cite{ pfeiffer2017perception} or simulation-based training \cite{loquercio2021learning}.
Reinforcement learning (RL) provides an alternative by enabling policies to explore the action space independently of expert labels. While RL benefits from exploration, it also encounters obstacles such as the sim-to-real gap and low sample efficiency \cite{ye2021auxiliary}. Techniques like auxiliary tasks \cite{zhu2020vision} have been proposed to address these limitations, while being sensitive to the accuracy of the data.


\subsection{Humanoid Robot Walking Planners}

Humanoid robots often require mid-level step planners to convert rough path plans into executable step sequences. Hildebrandt et al. \cite{hildebrandt2017real} combined mobile platform planners with step planners to enhance real-time planning, using sensor-based navigation to generate paths in dynamic environments. Xiong et al. \cite{xiong2021global} developed a step-to-step MPC based on a hybrid-linear inverted pendulum (H-LIP) model, optimizing footsteps while using feedback control to handle model approximation errors.
Control methods based on the divergent component of motion (DCM) \cite{khadiv2020walking} allow continuous adjustments of current stepping through closed-loop control. However, traditional methods typically separate high-level path planning from mid-level step planning and low-level control. Our work addresses this gap by integrating end-to-end planning and control from visual sensor inputs, employing task-level losses and combining collision maps with MPCs for physics-based guidance. To the best of our knowledge, iWalker is one of the first end-to-end learned vision-to-control solutions for humanoid robots walking.

\subsection{Imperative Learning and Bi-level Optimization}

Imperative learning (IL) is an emerging neural-symbolic learning framework based on bi-level optimization (BLO) \cite{wang2024imperative}.
It has been applied to various fields in robotics \cite{fu2024islam,zhan2024imatching,chen2024iastar} due to its enhanced accuracy offered by the symbolic reasoning engines and self-supervised learning nature.
For instance, iSLAM \cite{fu2024islam} formulated the simultaneous localization and mapping (SLAM) problem as a BLO, enabling the reciprocal optimization of front-end visual odometry and back-end pose graph optimization. 
iPlanner \cite{yang2023iplanner} employed a planning network to generate waypoints, followed by a closed-form solution for path interpolation in the lower-level optimization.
ViPlanner \cite{roth2024viplanner} introduced semantic information into the planning network, enabling the ability to distinguish between different terrains.
iKap \cite{li2025ikap} introduced kinematic constraints for generating waypoints, enabling kinematics-feasible planning for quadruped robots.

Bi-level optimizations and similar approaches also have applications on actual robot control and design. For example, \cite{amos2018differentiable} enables gradients to flow through the control optimization process, allowing the entire planning and control pipeline to be trained end-to-end based on task-specific loss functions. It allows upper-level networks to optimize with differentiable low-level MPCs, but we achieve a similar effect with non-differentiable MPCs;
\cite{fu2025anynav} used a similar manner to infuse frictional physics into robot control networks with BLO in training; \cite{sartore2023codesign}, interestingly, propose a bi-level optimization framework for humanoid robot design, integrating genetic algorithms for link and motor selection with nonlinear optimization to enhance ergonomic collaboration in payload lifting tasks, improving energy efficiency.

Further extending prior works, iWalker simultaneously incorporates two bi-level optimization loops that infuse humanoid dynamics models and MPCs into networks, predicting physically feasible solutions for walking humanoids.

\section{Methodology}

\begin{figure}
    \centering\includegraphics[width=\linewidth]{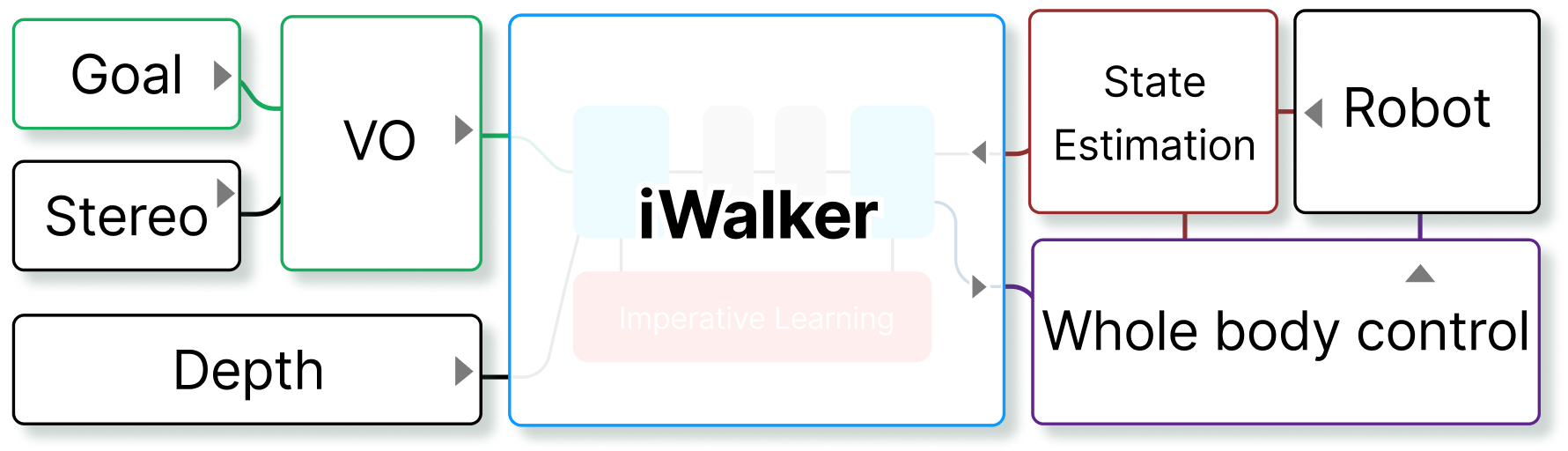}
    \caption{The system pipeline of our humanoid robot, where iWalker is a walking planner for mid-level footstep control.}
    \label{Fig. system}
\end{figure}

As shown in \fref{Fig. system}, iWalker serves as an end-to-end visual mid-level stepping controller, which connects sensors and commands to low-level control for a humanoid robot.

\subsection{Overall Structure of iWalker}

\begin{figure}[t]
\centering\includegraphics[width=0.5\textwidth]{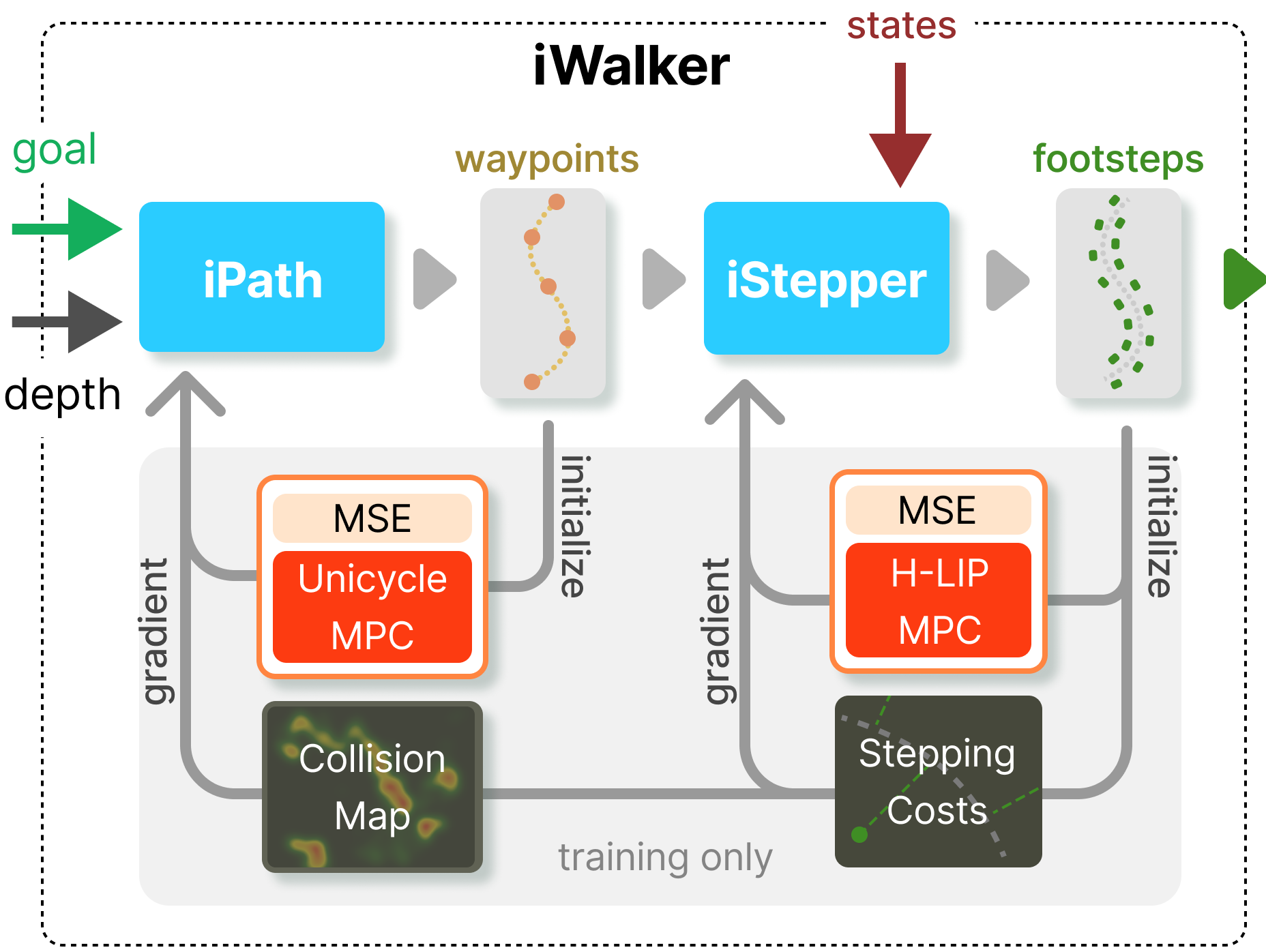}
\caption{iWalker includes two bi-level optimizations (BLO) based on imperative learning (IL). The first BLO optimizes iPath to predict a dynamically optimal and collision-safe path, while the second BLO optimizes iStepper to predict physically feasible footsteps for low-level controllers. iPath uses the same network structure as iPlanner, while iStepper is a simple three-layer MLP model.}
\label{Fig. iwalker}
\end{figure}

As illustrated in \fref{Fig. iwalker}, iWalker is a system consisting of two primary components, a visual path planner and a mid-level step controller, each accompanied by a bi-level optimization (BLO). The visual path planner, implemented as a neural network, processes depth images and goal positions to generate sequences of waypoints. Building on this, the step controller, another simple network, receives the path and the robot's state as inputs and outputs the next feasible footstep for low-level whole-body control.

This design offers several advantages for walking a humanoid robot. First, the system implements a complete sense-plan-act pipeline, eliminating the necessity for complex online map construction. Second, leveraging imperative learning \cite{wang2024imperative}, the system can be trained in a self-supervised end-to-end manner, which mitigates the compounding errors introduced by individual components. Third, during training, constraints from the robot's dynamic models are infused into the networks by solving bi-level optimizations, ensuring that the planned paths and footsteps are physically feasible for our chosen humanoid robot.
Since both the visual planner and the step controller are based on imperative learning (IL), we refer to them as ``iPath'' and ``iStepper'', respectively. We next introduce the details of the two modules in \sref{sec:ipath} and \sref{sec:istepper}, respectively. The methods of low-level whole-body control are presented in \sref{sec:WBC}.

\subsection{Dynamics-aware Path Planning (iPath)} \label{sec:ipath}

As a visual path planner, iPath takes a goal location and a depth image to generate paths, which guide the robot toward the target while avoiding obstacles. To infuse dynamics, we follow the imperative learning (IL) framework \cite{wang2024imperative} with physics-constrained MPCs to train iPath via a bi-level optimization (BLO).
Specifically, our goal is to train a network $f_{\text{path}}$ so that it outputs a trajectory that can be easily tracked and fitted by our chosen MPC, resulting in a minimum upper-level loss $U_{\text{path}}$ subjecting to a lower-level loss $L_{\text{path}}$.
The optimization schemes can be formulated as:
\label{eq:ipath}
\begin{align}\label{eq:ipath-ul}
&\min_{w_p} \;\; U_{\text{path}}(f_{\text{path}}(d, g; w_p), \phi^*), \\
&\; \textrm{s.t.} \;\; \phi^* = \arg\min_{\phi}~L_{\text{path}}(\hat{\phi}, \phi), \label{eq:ipath-ll}\\
&\;\;\;\;\;\;\;\;\hat{\phi} = f_{\text{path}}(d,g; w_p),\\
&\;\;\;\;\;\;\;\;U_{\text{path}} = \text{MSE}(\hat{\phi},\phi^*),
\end{align}
where $\hat{\phi}$ is the predicted trajectory from network $f_{\text{path}}$ with an optimized path $\phi^*$ as the optimization target, $w_p$ as weights, $d$ as depth inputs, and $g$ as the goal location. Specifically, the self-supervision of the upper-level network training \eqref{eq:ipath-ul} is achieved by enforcing the lower-level optimization in \eqref{eq:ipath-ll}, which is an MPC requiring no labels. The objective $L_\text{path}$ of the  lower-level MPC for $N$ states is:
\begin{align}
L_{\text{path}}(\hat{\phi}, \phi) &= \Xsum_{k}^{N} (\phi_k - \hat{\phi}_k)^T Q (\phi_k - \hat{\phi}_k) + u_k^T R u_k, \\
&\textrm{s.t.} \;\;\;\; \phi_{k+1} = \xi(\phi_k,u_k),
\end{align}
where $k$ is time index; the predicted trajectory $\hat{\phi}$ is used as reference for state $\phi$; $u$ is dynamic inputs; $\xi$ is the MPC transition function; and $Q$ and $R$ are weight matrices of quadratic costs. Intuitively, if $U_{\text{path}}$ and $L_{\text{path}}$ reach their minimum, which means the network's prediction is nearly optimized, we can say prediction $\hat{\phi}$ is physically feasible.

To optimize a smoother trajectory $\phi^*$, we choose a unicycle MPC which minimizes the lower-level loss $L_\text{path}$. By penalizing linear acceleration and angular velocity, we can improve the smoothness and evenness of the trajectory $\phi$. This is because a constant speed ensures evenly distributed waypoints and a low angular velocity minimizes turnings.
The discrete dynamics of the unicycle model is given by:
\begin{subequations}
\begin{align}
    x_{k+1} &= x_k + v_k \cos{\theta_k} \cdot dt,\\
    y_{k+1} &= y_k + v_k \sin{\theta_k} \cdot dt,\\
    v_{k+1} &= v_k + a_k \cdot dt,\\
    \theta_{k+1} &= \theta_{k} +\omega_k \cdot dt,
\end{align}
\end{subequations}
where we use $[x_k,y_k,v_k,\theta_k]^T$ as the state vector, representing the robot's position $x_k, y_k$, velocity $v_k$, and orientation $\theta_k$; $[ \omega_k, a_k ]^T$ is the control input, consisting of angular velocity $\omega_k$ and the acceleration $a_k$.
To solve this non-linear unicycle dynamics we use an iterative QP solver like iLQR \cite{li2004iterative} implemented by OSQP \cite{osqp}, and PyPose \cite{wang2023pypose} modules are used to realize BLO.
A low $U_{\text{path}}$ indicates that the predicted $\hat{\phi}$ is close to the optimized $\phi^*$, and a low $L_\text{path}$ means it is easy for MPC to track $\hat{\phi}$ given costs and constraints.

Although iPath shares the same network architecture with iPlanner \cite{yang2023iplanner}, our training approach is significantly different. iPlanner only relies on multiple heuristic losses like motion cost and goal cost based on waypoint locations without any dynamics constraints, while iPath, instead, uses one single unicycle MPC loss and demonstrates superior performance enhancement across all metrics, as shown in experiments.

\begin{algorithm}[t]
\caption{Optimization Iteration}\label{alg: optimize}
\textbf{Input:} depth $d$, goal $g$, state $\epsilon$\\
$\hat{\phi} \gets f_{\text{path}}(d,g)$\\
$\phi^* \gets g_{\text{uni-MPC}}(\hat{\phi})$\\
$U_{\text{path}} \gets \text{MSE}(\hat{\phi},\phi^*)$\\
\textbf{backward} $U_{\text{path}}$\\

\vspace{2pt}
\hrule
\vspace{2pt}

Initialize collision map $M_{\text{col}} \in \mathbb{Z}^{h \times w}$\\
$P \gets \text{project}(d)$\\
\ForEach{$p \in \text{3D point cloud }P$}{
    $(n, m) \gets \text{index}(p_x, p_y)$\\
    $M[n,m] \gets M[n,m] + 1$\\
}
$M_{\text{ESDF}} \gets G(M_{\text{col}})$\\

\vspace{2pt}
\hrule
\vspace{2pt}

$\hat{S} \gets f_{\text{step}}(\hat{\phi},\epsilon)$\\
$S^* \gets g_{\text{LIP-MPC}}(\hat{S},\epsilon)$\\
$U_\text{ESDF} \gets 0$\\
\ForEach{$\hat{s} \in \text{step sequence }\hat{S}$}{
    $U_\text{ESDF} \gets U_\text{ESDF} + \text{sample}(M_\text{ESDF},\hat{s_x},\hat{s_y})$ 
}
$U_{\text{step}} \gets \text{MSE}(S,S^*) + U_w(S,\Bar{w}) + U_l(S,\Bar{l}) + U_\text{ESDF}$\\
\textbf{backward} $U_{\text{step}}$\\
Optimize $f_{\text{path}}, f_{\text{step}}$\\
\Return
\end{algorithm}

\subsection{Dynamics-aware Mid-level Step Control (iStepper)}\label{sec:istepper}

After obtaining a raw path, we need to predict a series of footsteps considering the robot's state and dynamics. Similarly, we employ a second IL learning strategy for stepping predictor $f_\text{step}$, which can be formulated as:
\begin{align}
&\min_{w_s} \;\; U_{\text{step}}(f_{\text{step}}(\hat{\phi},\varepsilon; w_s), S^*), \\
&\; \textrm{s.t.} \;\; S^* = \arg\min_{S} L_{\text{step}}(\hat{S}, S), \\ 
& \;\;\;\;\;\;\;\;\hat{S} = f_{\text{step}}(\hat{\phi},\varepsilon; w_s),
\end{align}
where the network $f_\text{step}$ predicts a step sequence $\hat{S}$ given the previously predicted path $\hat{\phi}$ and current robot states $\varepsilon$. The network structure of iStepper as simple as a three-layer MLP, taking waypoints as input and outputing a series of 2D footstep locations, while only the first footstep will be executed by the low-level controller.

To make our prediction physically feasible for step control, we employ a step-to-step (S2S) dynamics-based MPC~\cite{xiong2021global}. Due to the difficulty of directly obtaining the analytical form of robot S2S dynamics, we use a linear approximation called the hybrid-linear inverted pendulum (H-LIP) model \cite{xiong20223} to approximate robot S2S dynamics, tracking the predicted footsteps. Specifically, the H-LIP dynamics is defined as:
\begin{equation}
    \mathbf{x}_{k+1}^\text{H-LIP} = A_s\mathbf{x}_{k}^\text{H-LIP} + B_su_k^\text{H-LIP},
\end{equation}
where $\mathbf{x}_{k}^\text{H-LIP} = [\mu_k, p_k, v_k]^T$ denotes the $k$-th pre-impact state of the H-LIP, with $\mu_k$, $p_k$, and $v_k$ being the global center of mass (COM) position, stance foot position w.r.t COM, and velocity, respectively; $u_k$ denotes the $k$-th global step size. The formulation of $A_s$ and $B_s$ can be found in \cite{xiong2021global}. 

\begin{figure}[t]
\centering\includegraphics[width=0.49\textwidth]{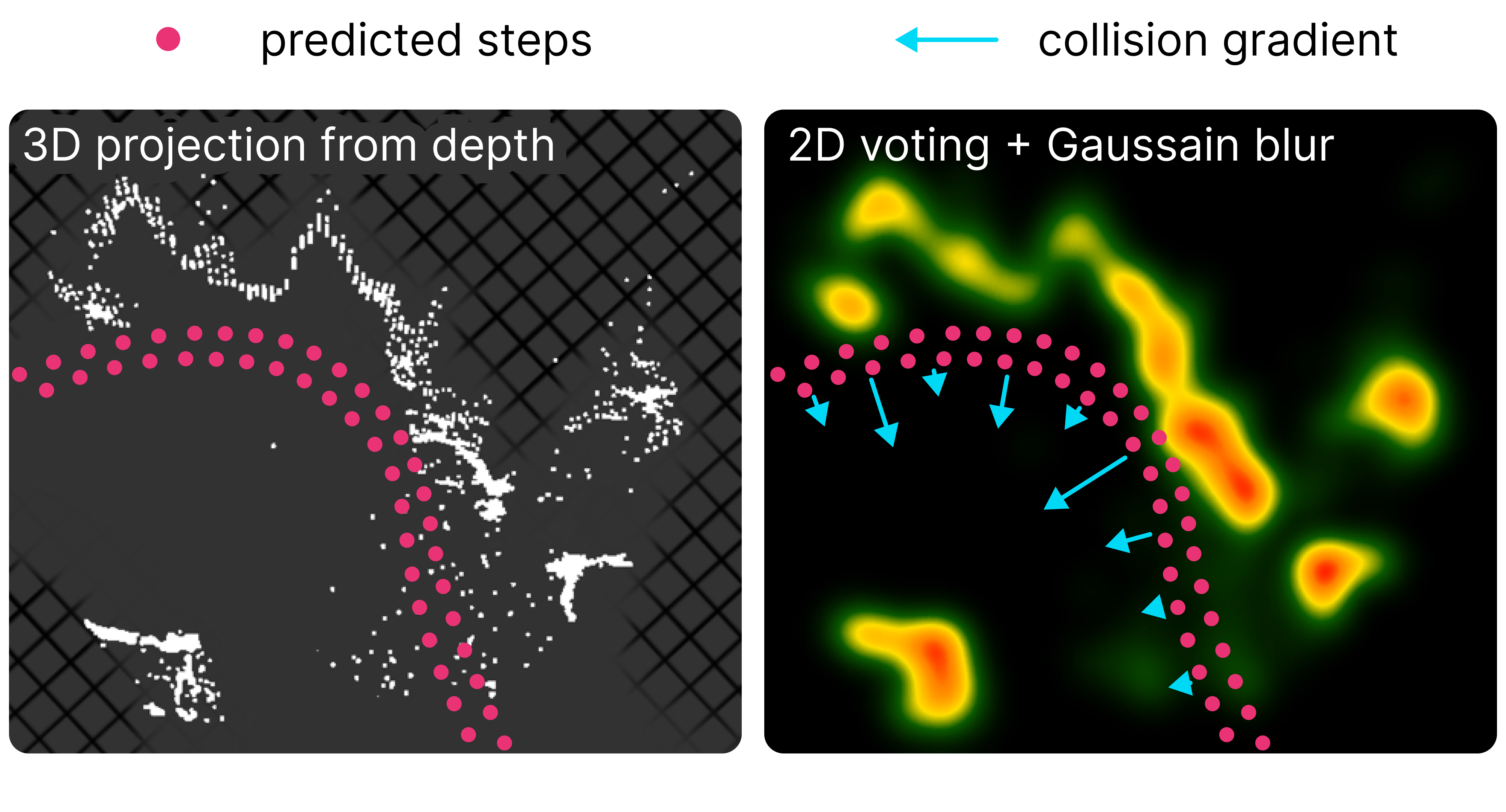}
\caption{A visualization of collision map and gradient. The 3D point cloud (left) is obtained from depth projection, while the right is an approximated ESDF map from blurring.}
\label{Fig. collision map}
\vspace{-20pt}
\end{figure}


Since the 2D H-LIP is based on two stacked 1D H-LIPs, modeling XY planes independently, it does not describe the robot's orientation, which essentially determines the kinematic feasibility of realizable step sizes. To encode the desired step width and step length in robot's frame to the step planner, we define two extra upper-level loss functions, $U_w$ on step width and $U_l$ on step length, with global-to-local mapping. These additional upper-level costs are measured in the robot's local frame given path $\hat{\phi}$, with $\Bar{w}$ and $\Bar{l}$ being the desired step width and step length, respectively:
\begin{align}
U_w &= \Xsum_k \text{MSE}(h_x(\hat{s_k},\hat{\phi}),\Bar{l})\;,\\
U_l &= \Xsum_k \text{MSE}(h_y(\hat{s_k},\hat{\phi}),\Bar{w}\cdot \alpha_k)\;,\\
\alpha_k &= 
\begin{cases} 
1, & \text{if } (k + o) \text{ is even}, \\
-1, & \text{if } (k + o) \text{ is odd},
\end{cases}
\end{align}
where $h_x$ and $h_y$ are functions that map one step location $s_k$ to its width and length in the robot's local frame, the origin and orientation are determined by the closest point on the path $\hat{\phi}$ from $\hat{s_k}$; To distinguish left-right stepping, $o$ is 0 if the robot is currently standing with the left leg, otherwise 1.

To enable obstacle avoidance capability, we build a Euclidean Signed Distance Field (ESDF) map $M_{\text{ESDF}}$ as described in the second block of Algorithms \ref{alg: optimize} and visualized in \fref{Fig. collision map}.
The ESDF map is used to calculate the collision costs and gradients subject to step locations, which is the key to ensuring collision-free stepping. 
Additionally, we apply a Gaussian blurring $G$ on the collision map to filter noise and ensure a continuous gradient space, guiding the footsteps to safer regions. We sample the collision costs by the stepping locations to ensure the feasibility of actual footsteps, instead of measuring directly from the raw waypoints.

The complete upper-level loss $U_{\text{step}}$ for step planning combining losses mentioned above is defined as:
\begin{equation}
\begin{aligned}
U_{\text{step}} &= \text{MSE}(\hat{S},S^*) + U_w + U_l + U_\text{ESDF}\;.\\
\end{aligned}
\end{equation}
This loss enforces the physical constraints described by the H-LIP, step targets, and collision costs. The corresponding part can be found in the second and third blocks in Algorithm \ref{alg: optimize}. For notation simplicity, we omitted the cost weights that balance multiple losses.
It is necessary to note that back-propagating the step planning loss $U_{\text{step}}$ produces gradients for both path planner $f_{\text{path}}$ and step controller $f_{\text{step}}$, which is the key to eliminate accumulated errors between modules by end-to-end learning. In contrast, path planning loss $U_{\text{path}}$ will only backward gradients for the weights in $f_{\text{path}}$. A detailed pseudo-code for one iteration of bi-level optimization based on imperative learning is listed in Algorithm \ref{alg: optimize}.

\subsection{Whole Body Control} \label{sec:WBC}

Once the desired footstep is determined, the next step is to execute it by a task-space whole-body controller \cite{shen2022locomotion}.
We formulate the controller in \eqref{eq:wbc}, where the tasks include tracking the swing foot trajectory, swing foot orientation, torso orientation, centroidal momentum, and contact forces.
\begin{subequations}\label{eq:wbc}
\begin{align}
    &\min_{\ddot{q}, \tau, F_j} \Xsum_{i=1}^{N_t} \left\| J_i \ddot{q} + \dot{J_i} \dot{q} - \ddot{x}_i^{\text{des}} \right\|^2_{W_i} 
    + \Xsum_{j=1}^{N_c} \left\| F_j \right\|^2_{W_f} + \left\|\ddot{q}\right\|^2_{W_{\ddot{q}}}, \\
    &~\text{s.t. } \quad H_b \ddot{q} + h_b(q,\dot{q}) = B\tau + \Xsum_{j=1}^{N_c} J_{j}^\top F_j, \\
    &\quad\quad~~A(F_j) \leq 0, \quad  \forall j = 1, \dots, N_c,
\end{align}
\end{subequations}
where $q$ denotes the robot configuration, $W$ denotes the weights on the cost, $J_i$ is the Jacobian of $i$-th task, $J_j$ is the $j$-th contact Jacobian, $N_t$ is the number of tasks, $F_j$ is the contact reaction force at the foot-ground contact indexed by $j$, $N_c$ is the number of contact points, $A(F_j)\leq 0$ denotes the linearized contact force constraints, $H_b$ is the mass matrix, $h_b$ denotes the Coriolis and gravity vector, $\tau$ is the motor torques to solve for execution, and $B$ is constant matrix mapping the motor torques to generalized forces on the robot joints. As we can see, the $i$-th task space goal is set as a cost in the quadratic program with priority implicitly being enforced with weight $W_i$. Additionally, regularization costs are added to the decision variables $\ddot{q}$ and $F_j$ with small weights $W_{\ddot{q}}$ and $W_f$, respectively. To effectively follow our first planned step location $\hat{s_0}$ from iWalker, a reference polynomial trajectory is fitted according to the current swinging foot's state.
\subsection{End-to-end Training}

Our training framework jointly optimizes iPath, iStepper, and the MPCs in an end-to-end manner. The total loss for iWalker networks is defined as:
\begin{equation}
U_{\text{total}} = U_{\text{path}} + U_{\text{step}},
\end{equation}
where $U_{\text{path}}$ exclusively optimizes the iPath module and $U_{\text{step}}$ influences both iPath and iStepper, enabling full differentiability with respect to the vision input. Training is performed with a batch size of 1 using the Adam optimizer. Data augmentation is applied by horizontally flipping both the depth image and the corresponding goal position. Our dataset comprises 1185 frames collected from 12 different indoor scenes using the RealSense camera. Each frame contains monochrome and depth image and the goal position typically set at the end of the sequence. During training, the MPCs are run after the predictions of iPath and iStepper to optimize low-level losses and generate high-level losses. Finally, a single backward pass is performed, ensuring that the entire pipeline benefits from end-to-end gradient flow.

\subsection{State Estimation}
State estimation is fundamental for both step planning and control, as it provides the necessary high-frequency pose and velocity information. To achieve this, we fuse sensor data from both the IMU and stereo camera. Specifically, the stereo images are processed by AirVO \cite{xu2023airvo}, which operates at 10-20Hz and estimates the robot's global position and orientation. Simultaneously, the IMU, coupled with a modified Kalman Filter (KF), provides high-frequency state updates of the torso velocity, primarily used for WBC.


\section{Experiments}


\subsection{Robot Platform and Software Architecture}
We evaluate our algorithms on BRUCE \cite{shen2022locomotion}, a 0.6~meter tall humanoid robot: its leg is equipped with five degrees of freedom (DoF) joint with proprioceptive actuators, and foot contact switches. Additionally, it has an Inertial Measurement Unit (IMU) located on the torso to facilitate sensing. We mounted a RealSense D455 camera (resolution of $640\times320$) on the top of the robot for depth perception. Since the onboard computer has limited computation capability, we connect the robot to a laptop-level computer that has an AMD 6900HS CPU and an NVIDIA 3070Ti GPU which send commands to motor drivers implemented by the vendor.  

Our software architecture adopts a multi-process design wherein a primary Python process manages overall system inputs and outputs while coordinating specialized modules. These modules include a Kalman filter \cite{shen2022locomotion} for sensor fusion, a controller module for mid-level trajectory generation and low-level motor actuation, a visual odometry based on \cite{xu2023airvo} for accurate localization, and our iWalker module that realizes vision-to-control framework for dynamic navigation. All processes are running on Ubuntu 20.04 using ROS1 \cite{quigley2009ros} environment. In our experiments, the iPath module operates at 15\,Hz, continuously generating path predictions. The iStepper module is updated at step-level at 4\,Hz, which plans desired subsequent footsteps based on the latest path prediction. The MPCs for BLOs are only used during training to supervise the predictions from iPath and iStepper.

\subsection{Baselines}

To the best of our knowledge, vision-to-control solutions for humanoid robots are rarely investigated in the literature. Also, open-sourced implementations of known algorithms are very limited and incompatible to our BRUCE robot platform. Thus, we choose iPlanner as our primary baseline, which shares the same network structure of our iPath module. We consider two configurations to provide comprehensive evaluations. One configuration, referred to as iPlanner$^*$, represents iPlanner fine-tuned solely using $U_{\text{path}}$ without the joint training with iStepper. In the final configuration for iWalker, iStepper and iPath are jointly trained to evaluate the effectiveness of our proposed fully end-to-end training. 

Additionally, we explored a Reinforcement Learning (RL) approach using Proximal Policy Optimization (PPO) \cite{schulman2017proximal} with the MPC cost serving as the reward signal. However, this RL approach struggled to converge or improve, likely due to the complexity of our task. This outcome further underscores the importance of direct gradient computed from our MPC-based BLO technique for achieving robust and stable training, which is inspired by the IL framework.

\subsection{Quantitative Comparison}
We evaluated the performance of our methods and baselines using the following metrics on four recorded test scene sequences: \textit{Stair}, \textit{Office}, \textit{Corridor}, and \textit{Corner}.

\subsubsection{The Dynamical Feasibility}
To measure the dynamical feasibility of predicted steps, we measure the low-level costs from unicycle and H-LIP MPC. Intuitively, lower MPC costs indicate less violation of the robot's dynamics and better performance. As shown in \tref{tab:path}, iWalker, trained iPath and iStepper in an end-to-end manner, eliminates the compound errors and achieves significant improvement, while both iPlanner and iPlanner$^*$ are unaware of dynamics needed for low-level control and thus have higher cost values. 

\subsubsection{Collision Safety}

To assess collision safety, we calculate the collision risk by sampling the collision ESDF maps at the predicted footstep locations. Intuitively, lower risk values correspond to safer predictions, indicating improved obstacle avoidance along the step sequence. As shown in \ref{tab:collision}, iWalker has the lowest collision risk on the predicted steps, while both iPlanner and iPlanner only penalize collision costs at waypoints, leaving footstep-level safety unaddressed.

\subsubsection{Waypoints Stability} 

To measure the stability of predicted waypoints before step control, we measure the variance of the waypoint interval as an indicator of waypoint evenness. This is because evenly distributed waypoints can lead to easier MPC optimization and smoother low-level actuation. As shown in \tref{tab:evenness}, iPlanner, which was trained on accurate LiDAR maps, fails to generate stable waypoints given our noisy depth sensing, no matter whether fine-tuned or not. In contrast, iWalker consistently improves waypoint evenness by incorporating dynamic information.



\begin{table}[!t]
    \caption{The violation of dynamical feasibility ($\downarrow$).}
    \label{tab:path}
    \centering
    \resizebox{\linewidth}{!}{
    \begin{tabular}{ccccc}
      \toprule
      Method & Stair & Office & Corridor & Corner\\
      \midrule
        iPlanner + iStepper    & 1.4840 & 2.4366 & 1.286 & 0.8140 \\
        iPlanner$^*$+iStepper & 3.5751 & 2.2521 & 5.2924 &  5.7913\\
        \textbf{iPath + iStepper}                & \textbf{1.4321}& \textbf{0.4827} & \textbf{0.4677} & \textbf{0.5330}\\
      \bottomrule
    \end{tabular}
    }
    \vspace{0.4em} 
    \caption{The evalutation of collision safety ($\downarrow$).}
    \label{tab:collision}
    \centering
    \resizebox{\linewidth}{!}{
    \begin{tabular}{cccccc}
      \toprule
      Method & Stair & Office & Corridor & Corner \\
      \midrule
        iPlanner + iStepper    & 11.969 & 21.522     & 4.4698 & 5.8537 \\
        iPlanner$^*$ + iStepper & 12.425 & 24.905    & 3.8941 & 6.5459 \\
        \textbf{iPath + iStepper}                  & \textbf{10.167} & \textbf{8.9221}     & \textbf{3.1317} & \textbf{3.998} \\
      \bottomrule
    \end{tabular}
    }
    \vspace{0.4em} 
    \caption{The performance of waypoint stability ($\downarrow$).}
    \label{tab:evenness}
    \centering
    \resizebox{\linewidth}{!}{
    \begin{tabular}{ccccc}
      \toprule
      Method & Stair & Office & Corridor & Corner\\
      \midrule
        iPlanner + iStepper    & 89.8& 3.20 & 19.1 & 1.00\\
        iPlanner$^*$ + iStepper & 71.6 & 24.6 & 33.3 & 38.3\\
        \textbf{iPath + iStepper}                  & \textbf{31.0} & \textbf{0.90}     & \textbf{3.32} & \textbf{0.71} \\
      \bottomrule
    \end{tabular}
    }
    \vspace{0.6em} 
    \caption*{Note: Lower values indicate better performance across all metrics. We implemented PPO RL \cite{schulman2017proximal} as a baseline, but it failed to produce consistent results, which is thus excluded from the table. We denote iWalker as ``iPath + iStepper'' for easier interpretation as they are equivalent. All iSteppers are the same and jointly trained with iPath.}
\end{table}

\begin{figure*}[t]
\centering\includegraphics[width=1\textwidth]{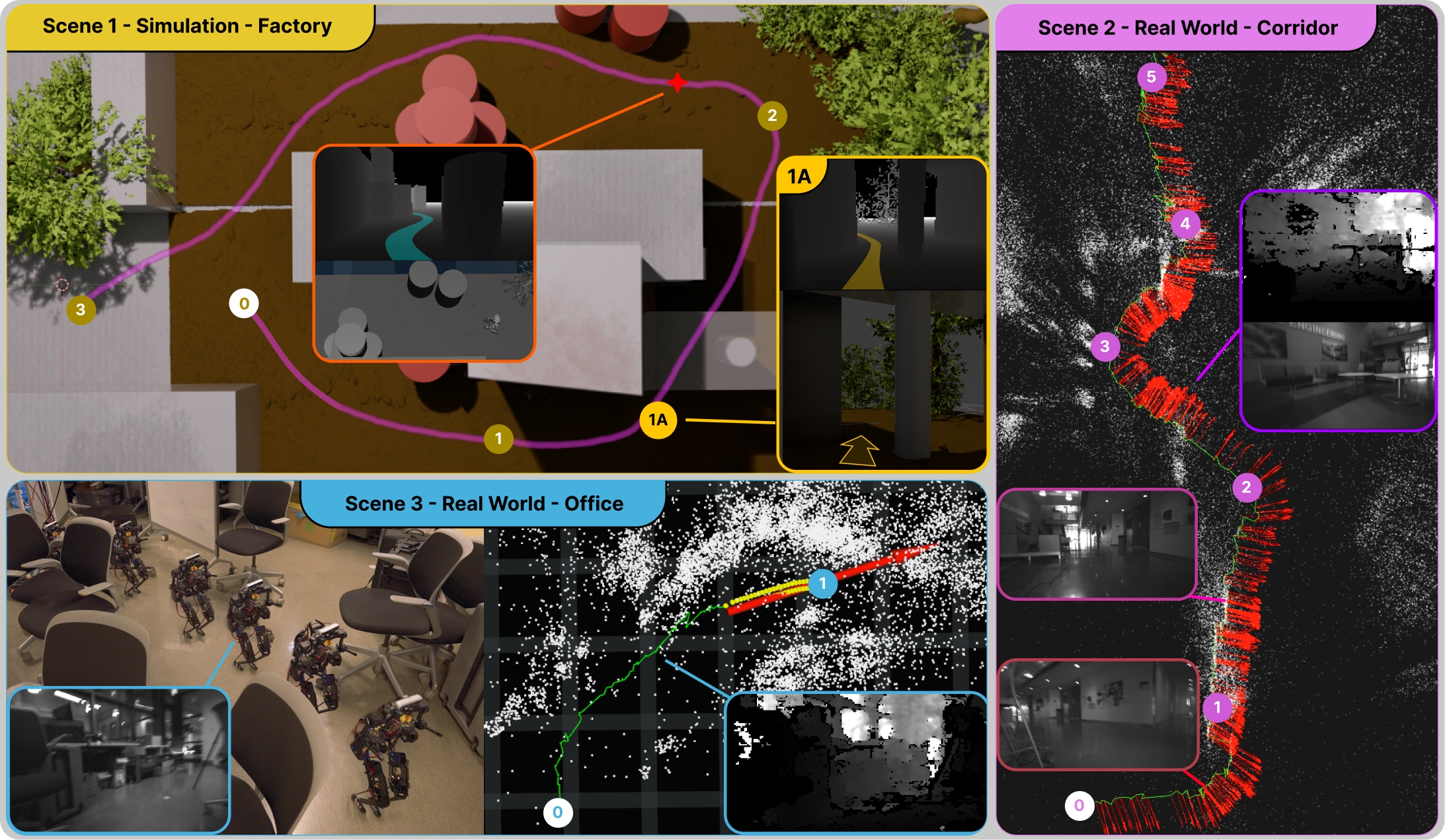}
\caption{We presents both simulation and real-world demonstrations of our robot's navigation capabilities across distinct environments. Goal points are explicitly labeled. In \textbf{Scene 1}, the robot with iWalker trained on real-world data navigates in a factory-like scene using only three goal points, showcasing efficient navigation in unseen environments. In \textbf{Scene 2}, the robot traverses a 10-meter corridor with many turns, showing its proficiency in handling long pathways. In \textbf{Scene 3}, iWalker navigates a short path through environments with irregular shapes, illustrating its adaptability to noisy depth.}
\label{Fig. demo}
\end{figure*}
\subsection{Real-time Evaluation in Simulation}

We primarily tested our complete sense-plan-act pipeline for the humanoid robot BRUCE with iWalker within the MuJoCo simulation environment \cite{todorov2012mujoco}.
It is worth noting that in this experiment, iWalker is only trained with real-world noisy depth images, without seeing any images from simulation.
In \fref{Fig. demo}-Scene 1, the robot smoothly navigates through a large factory environment. This demonstrates that iWalker can effectively adapt to unseen simulated environments non-trivially different from real-world training data. iWalker can complete fully autonomous navigation across half of the map from point 2 to point 3 under one single goal command, traversing through the barrels and unseen obstacles. Moreover, iWalker is able to navigate through a narrow gap between the wall and the pillar without collision as highlighted in \fref{Fig. demo}-1A. With stable low-level control and precise state estimation in the simulation, iWalker can always plan and walk smoothly even in unseen scenes.

\subsection{Real-time Demo with Real Humanoid Robot}

We next evaluate iWalker's performance on the real robot. We command the robot to navigate through a long path in \fref{Fig. demo}-Scene 2, where the robot traversed a long 10-meter corridor with multiple turns. This verifies the effectiveness of iWalker in large-scale environments. In addition, we tested iWalker to pass through messy environments such as \fref{Fig. demo}-Scene 3, where iWalker successfully navigates through a narrow path in a cluttered office with randomly placed chairs.

These tests demonstrate iWalker's generalization capability and consistent performance in unseen and structured simulation scenes as well as noisy real-world environments, which validates its effectiveness across a variety of scenes. A video of the real-time demonstration is available at \url{https://sairlab.org/iwalker/}, in which we test iWalker under a more diverse set of settings and environments.

\section{Conclusion \& Limitations}

We introduced iWalker, a vision-to-control pipeline developed for humanoid robot walking. By incorporating two MPC-based bilevel optimizations through imperative learning, iWalker enables efficient and robust self-supervised training on visual data, predicting \textit{dynamically optimal and collision-free paths} and \textit{physically feasible footsteps} for low-level control. Our experiments, conducted in simulation and on real robots, demonstrate the capability of iWalker to navigate through various simulated and real-world environments. 

The future work will be on addressing the limitations of the current system design. Particularly, the low-level optimization-based whole-body controllers can yield infeasibility under extreme maneuvers, perhaps due to the limited capability of our small robot platform. This restricts our evaluation of the current work on more challenging environments. Thus, in the next, we will focus on integrating the feasibility of the low-level control, to further enhance the performance of the proposed vision-to-control framework. 

\section*{Acknowledgements}

This work was supported by the DARPA award HR00112490426. Any opinions, findings, conclusions, or recommendations expressed in this paper are those of the authors and do not necessarily reflect the views of DARPA.

\balance

\bibliographystyle{./bib/IEEEtran}
\bibliography{root,sairlab}

\end{document}